%% file: main.tex
\pdfoutput=1

\documentclass[11pt]{article}
\input{nabra_packages}
\input{nabra_macros}

\input{nabra_title_author}

\begin{document}
\setcode{utf8}
\maketitle
\input{nabra_abstract}

\section{Introduction}
\label{s:intro}
\input{nabra_s1_intro}

\section{Related work}
\label{s:related}
\input{nabra_s2_related}

\section{Syrian as a Levantine Dialect}
\label{s:syrian}
\input{nabra_s3_syrian}

\section{Variant Syrian Dialects}
\label{s:local}
\input{nabra_s4_local}

\section{\nabra Corpora Collection}
\label{s:collection}
\input{nabra_s5_collection}

\section{Annotation Methodology and Features}
\label{s:annotation}
\input{nabra_s6_annotation}

\section{Evaluation and Agreement} 
\label{s:eval}
\input{nabra_s7_evaluation}

\subsection{Qualitative Evaluation} 
\label{s:Qual_eval}
\input{nabra_s8_qualitative_evaluation}

\vspace{-.5em}

\section{Conclusion} 
\label{s:conclusion}
\vspace{-.5em}

This paper presents \nabra, a morphologically annotated corpora of Syrian Arabic dialects. 
The corpora contain about $60K$ tokens from $10$ Syrian dialects, collected from social media platforms, movies and series, common proverbs, and song lyrics and poetry. 
To be compatible with SAMA and other Arabic corpora, we chose to annotate the corpora using SAMA tagsets.
To evaluate the quality of the corpora, we used the $F1$ and $kappa$ scores which show high agreement. 

We plan to use \nabra to extend Wojood \cite{JKG22,LJKOA23} by annotating the corpora for Named Entity Recognition, similar to what we did with Curras and Baladi.

\section*{Limitations}
\label{s:limits}
\input{nabra_limitations}

\section*{Ethics Statement}

The collection of texts used in 
\nabra respects intellectual
property of the material. 
The annotation process employed annotators who were paid a fair rate per hour based on their living locality.
Results from \nabra will be shared online for the research community to use and improve upon. 

\section*{Acknowledgements}
We would like to thank all the students who helped in the annotation process, especially Salam Eideh, Nagham Idrees, Shahd Khader, Shimaa Hamayel, Rwaa Assi, Hiba Zaid, Lama Saqqour, Ali Maihoob, and Eman Hariri.

\bibliography{MyReferences,nabra_refs,nabra_refs_jarrar_zaraket}
\bibliographystyle{acl_natbib}

\appendix

\section{Appendix: \nabra Statistics}
\label{sec:appendix}

\begin{table}[h!]
    \centering
    \caption{Distribution of Gender feature. Arabic Words especially verbs and nouns and some of the functional words are annotated with ``Male''``Female''. In some cases, the gender can be both, depending on the context, such as \TrAr{الجميع} (everyone).}
    \begin{tabular}{l|l}
    \hline
        \bf{Gender} & \bf{Count} \\ \hline
        Male & 25,538 \\ 
        Female & 11,790 \\ 
        Both & 931 \\  \hline
    \end{tabular}
    \label{ref:gender_table}
\end{table}

\begin{table}[h!]
    \centering
    \caption{Distribution of the Number feature. Arabic words especially verbs and nouns are annotated with ``Singular'', ``Dual'', ``Plural'', and in some rare cases, the number can be "Any" like \TrAr{أبدى} (more important).} 
    \begin{tabular}{l|l}
    \hline
        \bf{Number} & \bf{Count} \\ \hline
        Singular & 32,372 \\
        Dual & 192 \\ 
        Plural & 4,450 \\  
        Any & 163 \\ \hline
    \end{tabular}
    \label{ref:number_table}
\end{table}

\begin{table}[h!]
    \centering
    \caption{Distribution of the verbs' Person: 1st person (\Ar{متكلم}), 2nd person (\Ar{مخاطب}), 3rd person (\Ar{غائب}). }
    \begin{tabular}{l|l}
    \hline
        \bf{Person} & \bf{Count} \\ \hline
        1st & 2,767 \\ 
        2nd & 2,794 \\
        3rd & 6,769 \\ \hline
    \end{tabular}
    \label{ref:person_table}
\end{table}

\begin{table}[ht!]
\small 
\centering
\caption{Distribution of the POS tags and categories. }
\label{ref:pos_table}
\begin{tabular}{|l|l|r|}
\hline
\bf{Category} & \bf{POS} & \bf{Count} \\ \hline
\multirow{9}{*}[-1.5ex]{\text{\begin{tabular}[c]{@{}c@{}}\textbf{NOUN}\\ \\ Total: 28,932\end{tabular}}} & NOUN & 21,250 \\ \cline{2-3} 
                             & ADJ              & 4,742  \\ \cline{2-3} 
                             & NOUN\_PROP       & 1,540  \\ \cline{2-3} 
                             & NOUN\_QUANT      & 556   \\ \cline{2-3} 
                             & NOUN\_NUM        & 315   \\ \cline{2-3} 
                             & ADJ\_COMP        & 257   \\ \cline{2-3} 
                             & ADJ\_NUM         & 152   \\ \cline{2-3} 
                             & ABBREV           & 31    \\ \cline{2-3} 
                             & DIGIT *         & 89    \\ \hline
 \multirow{5}{*}[-1.5ex]{\text{\begin{tabular}[c]{@{}c@{}}\textbf{VERB}\\ \\ Total: 11,166\end{tabular}}}
& IV               & 5,926  \\ \cline{2-3} 
                             & PV               & 3,846  \\ \cline{2-3} 
                             & CV               & 1,080  \\ \cline{2-3} 
                             & IV\_PASS         & 289   \\ \cline{2-3} 
                             & PV\_PASS         & 25    \\ \hline
\multirow{28}{*}[-1.5ex]{\text{\begin{tabular}[c]{@{}c@{}}\textbf{FUNC\_WORD} \\ \\ Total: 19,923\end{tabular}}} & PUNC *           & 5,010  \\ \cline{2-3} 
                             & PREP             & 3,133  \\ \cline{2-3} 
                             & CONJ             & 2,506  \\ \cline{2-3} 
                             & NEG\_PART        & 1,642  \\ \cline{2-3} 
                             & ADV              & 1,485  \\ \cline{2-3} 
                             & PRON             & 1,252  \\ \cline{2-3} 
                             & SUB\_CONJ        & 991  \\ \cline{2-3} 
                             & REL\_PRON        & 687   \\ \cline{2-3} 
                             & DEM\_PRON        & 645   \\ \cline{2-3} 
                             & INTERROG\_PART   & 489   \\ \cline{2-3} 
                             & VOC\_PART        & 357   \\ \cline{2-3} 
                             & PART             & 342   \\ \cline{2-3} 
                             & PROG\_PART *     & 218   \\ 
                             \cline{2-3} 
                             & VERB             & 171   \\ 
                             \cline{2-3} 
                             & INTERROG\_PRON   & 166   \\ \cline{2-3} 
                             & FUT\_PART        & 130   \\ \cline{2-3} 
                             & RESTRIC\_PART    & 117   \\ 
                             \cline{2-3} 
                             & FOREIGN          & 115   \\ \cline{2-3} 
                             & PSEUDO\_VERB     & 101   \\ 
                             \cline{2-3} 
                             & EMOJI *          & 95  \\ 
                             \cline{2-3} 
                             & VERB\_PART       & 44    \\ 
                             \cline{2-3} 
                             & INTERJ           & 43   \\ \cline{2-3} 
                             & DET              & 40    \\ \cline{2-3} 
                             & INTERROG\_ADV    & 38    \\ \cline{2-3} 
                             & EXCLAM\_PRON     & 35    \\ \cline{2-3} 
                             & FOCUS\_PART      & 33    \\ \cline{2-3} 
                             & PREP + SUB\_CONJ & 27    \\ \cline{2-3} 
                             & REL\_ADV         & 11    \\ \hline \hline
                              &\textbf{Total}         & \textbf{60,021}    \\ \hline
\end{tabular}
\end{table}

\end{document}

%% file: nabra_packages.tex
\usepackage{EMNLP2023}

\usepackage{times}
\usepackage{latexsym}

\usepackage[T1]{fontenc}

\usepackage[utf8]{inputenc}

\usepackage{microtype}

\usepackage{inconsolata}

\usepackage{amsmath,amssymb,amsfonts}
\usepackage{algorithmic}
\usepackage{graphicx}
\usepackage{textcomp}

\usepackage{arabtex}
\usepackage{utf8}
\usepackage{url}
\usepackage{xspace}
\usepackage{xcolor}
\usepackage{makecell}
\usepackage[colorinlistoftodos,prependcaption,textsize=tiny]{todonotes}

\usepackage{tipa} 
\usepackage{enumitem}
\setitemize{noitemsep,topsep=0pt,parsep=0pt,partopsep=0pt}

\usepackage{multirow}
 
\usepackage{abstract}

%% file: nabra_macros.tex
\def\BibTeX{{\rm B\kern-.05em{\sc i\kern-.025em b}\kern-.08em
    T\kern-.1667em\lower.7ex\hbox{E}\kern-.125emX}}

\newcommand{\Ar}[1]{{\small \<#1>\xspace}}

\newcommand{\TrNoAr}[1]{\arabfalse\transtrue \RL{\Ar{#1}}\arabtrue\transfalse}

\newcommand{\TrAr}[1]{\arabtrue\transfalse {\scriptsize \Ar{#1}} /\arabfalse\transtrue \RL{#1}\arabtrue\transfalse} 

\newcommand{\nabra}{\cal{N\^{a}b\=ra}\xspace}
\newcommand{\arnabra}{\Ar{نَبْرَة}\xspace}

%% file: nabra_title_author.tex
\title{Nabra: Syrian Arabic Dialects with Morphological Annotations}

\author{Amal Nayouf \\ Syrian Virtual University, Syria\\ \texttt{amal\_124724@svuonline.org}~~~~~ \\
\And
  {\bf Tymaa Hasanain Hammouda} \\
  Birzeit University, Palestine  \\
 ~~~~~\texttt{thammouda@birzeit.edu} \\ \And
  Mustafa Jarrar \\
  Birzeit University, Palestine  \\
\texttt{mjarrar@birzeit.edu} \\ \AND
  Fadi A. Zaraket, \texttt{zfadi@utexas.edu} \\
  Doha Institute for Graduate Studies, Doha \\
  American University of Beirut, Beirut 
   \\ \And
  Mohamad-Bassam Kurdy \\
  Syrian Virtual University, Syria \\
  \texttt{t\_bkurdy@svuonline.org} \\
}

%% file: nabra_abstract.tex
\begin{abstract}
This paper presents \nabra~(\arnabra), a corpora of Syrian Arabic dialects with morphological annotations. 
A team of Syrian natives collected more than $6K$ sentences containing about $60K$ words from several sources including social media posts, scripts of movies and series, lyrics of songs and local proverbs to build \nabra.
\nabra covers several local Syrian dialects including those of 
 Aleppo,
 Damascus,
 Deir-ezzur,
 Hama,
 Homs,
 Huran,
 Latakia,
 Mardin,
 Raqqah,
and
 Suwayda
.
A team of nine annotators annotated the $60K$ tokens with full morphological annotations across sentence contexts.
We trained the annotators to follow methodological annotation guidelines to ensure unique morpheme annotations,
and normalized the annotations.
F1 and $\kappa$ agreement scores ranged between $74\%$ and $98\%$ across features,  
showing the excellent quality of \nabra annotations.  
Our corpora are open-source and publicly available as part of the Currasat portal \url{https://sina.birzeit.edu/currasat}.
\end{abstract}

%% file: nabra_s1_intro.tex
Dialectal Arabic (DA) content dominates 
informal writings in emails, social media, blogs, and social messaging. 
Interest in building computational resources for Arabic dialects 
has been in the rise to provide both (i) annotated corpora~\cite{jarrar2022lisan,SevenCorpora,khalifa-etal-2018-morphologically,bouamor-etal-2018-madar,JHRAZ17,al-shargi-etal-2016-morphologically,ZribiEBB15,JHAZ14}
and (ii) morphological dialect analyzers~\cite{obeid-etal-2020-camel,khalifa-etal-2020-morphological,pasha-etal-2014-madamira,ZRIBI2017147,abdul-mageed-etal-2021-nadi}.  

In this paper, we present \nabra~\arnabra, a set of corpora that complement 
existing Arabic dialect corpora by covering several dialect variants of Syrian Arabic. 
\nabra covers dialects from $10$ Syrian localities including 
 Aleppo,
 Damascus (a.k.a. Shami) ,
 Deir-ezzur,
 Hama,
 Homs,
 Huran,
 Latakia,
 Mardin,
 Raqqah,
and
 Suwayda. 
\nabra was collected from several sources including 
social media posts, scripts of movies and series, lyrics of songs, and local proverbs.
Nine annotators worked on annotating 6K sentences with 60,021 tokens 
with full morphological annotations.
Each word was annotated using: prefix(s), stem, and suffix(s), part of speech (POS), dialect lemma, MSA lemma, person, number, gender, gloss, and synonyms; in addition to the sub-dialect it belongs to.

\begin{figure}[h]
    \centering
    \includegraphics[width=0.5\textwidth]{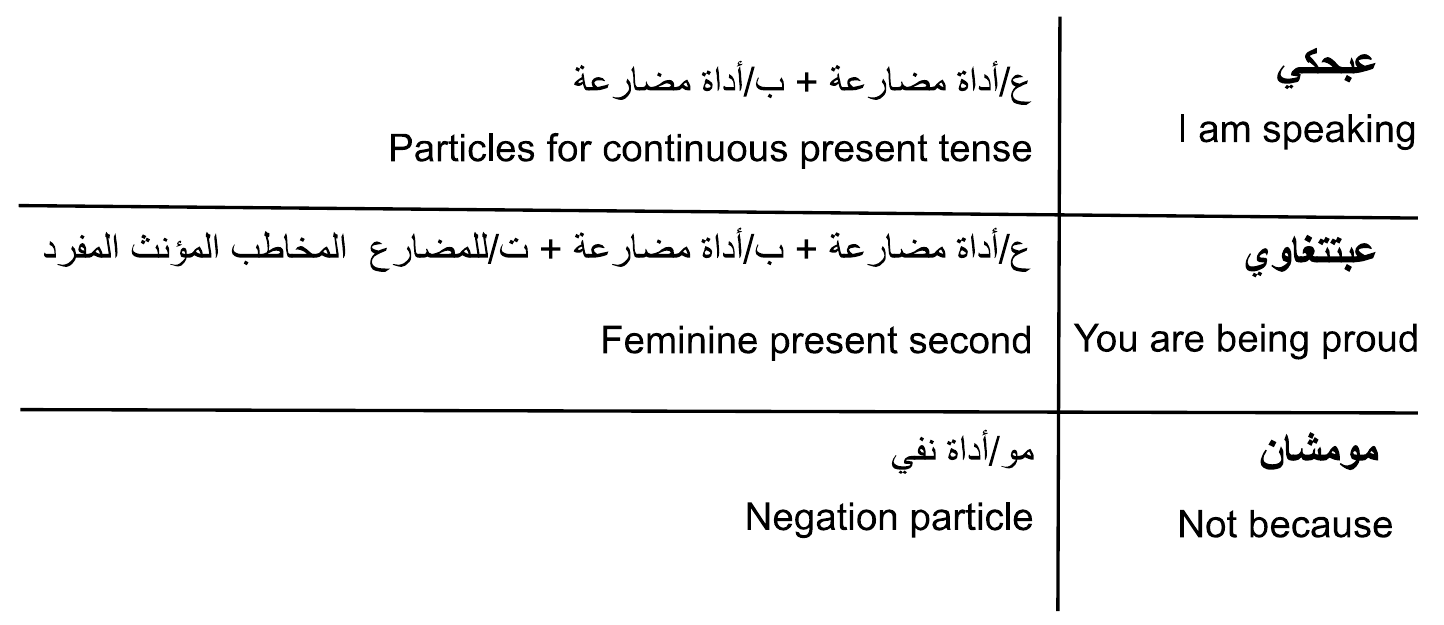}
    \caption{Examples of typical prefixes in Syrian dialects}
    \label{fig:syr_prefixes}
\end{figure}

\begin{figure}[h]
    \centering
    \includegraphics[width=0.5\textwidth]{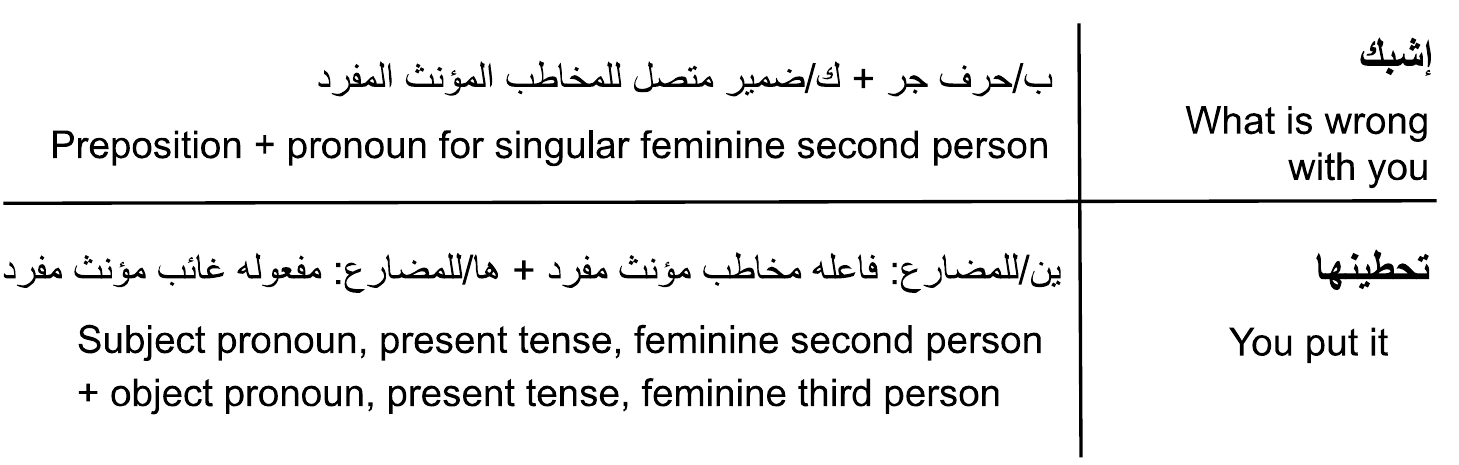}
    \caption{Examples of typical suffixes in Syrian dialects.}
    \label{fig:syr_suffixes}
\end{figure}

We adopted the same annotation methodology used to annotate the Palestinian Curras2 and the Lebanese Baladi corpora~\cite{EJHZ22}, which we also used with the four corpora of Lisan~\cite{JZHNW23}. 
As we will discuss later, we adopted the SAMA tagsets~\cite{MaamouriSama2010}, but we introduced new prefixes and suffixes that are commonly used in Syrian
dialects (Figures~\ref{fig:syr_prefixes}
and~\ref{fig:syr_suffixes}).

\subsection{Arabic and its Dialects}

Over 300 million people speak Arabic, including Classical Arabic (CA), Modern Standard Arabic (MSA), and dialectal forms of Arabic (DA), in more than $23$ countries. Natural language processing (NLP) research has traditionally focused on MSA because it is the most widely used form of Arabic in formal communication, newspapers, education, and media. CA dominates historical and cultural texts, whereas most colloquial and real-life communication uses local DA variants. 
DA content is lately gaining massive growth especially
through blogs, social media, and local entertainment outlets in songs, movies, and series. 

NLP pipelines often struggle with tasks involving DA content due to the inherent morphological richness of DA variants, their relative lack of resources compared to MSA, and the absence of a standardized orthography~\cite{DH21}. 
DA is classified regionally into Egyptian, Gulf, Levantine, North African, and Yemeni \cite{Colaba_diab_2010} 
with Syrian and Lebanese dialects considered as Northern Levantine, and Palestinian and Jordanian as Southern Levantine. 

Syrian Arabic 
is well-understood across the Arab world 
due to its popularity in
historical dramas, TV series, and soap operas. Twenty million Syrians speak it for daily life.
Expatriates from the Levant (Jordan, Lebanon, Palestine, and Syria) 
helped spread the dialect throughout the world.






The rest of this paper is organized as follows. Section~\ref{s:related} reviews related work.
We introduce Syrian as a Levantine dialect in Section~\ref{s:syrian} and discuss variant Syrian dialects in Section~\ref{s:local}.
\nabra data collection and annotation methodology
follow in Sections~\ref{s:collection} and~\ref{s:annotation}, respectively.
We discuss the evaluation of \nabra in Section~\ref{s:eval}, then we conclude 
in~\ref{s:conclusion} and discuss limitations and ethics considerations.

%% file: nabra_s2_related.tex
There are several annotated corpora and lexicographic resources for MSA.

The LDC's Penn Arabic Treebank PATB~\cite{MaamouriPATB2005} consists of about consists of 791,210 tokens collected from several news sources. PATB annotations include: tokenization, segmentation, POS tagging, lemmatization, diacritization, English gloss and syntactic structure. The LDC Ontonotes 5 \citep{ontonotes} is another MSA corpus collected from news sources, consisting of about 330K tokens, which are annotated in the same way as the PATB. Ontonotes 5 also contains multiple layers of annotation, including the PATB annotation layer.

The Prague Arabic Dependency Treebank (Ar-PADT)~\cite{PragueADT_2004} is a treebank that contains morphological annotations for a corpus of MSA text. These annotations include lemmas, part-of-speech tags, and other morphological features. Ar-PADT contains about 224K words.

The LDC's SAMA is a stem database \cite{MaamouriSama2010}, which is an extension of BAMA \cite{BuckwalterBAMA2004}, designed only for morphological modeling. It contains stems and their lemmas and compatible affixes. It contains about 40K lemmas.  

 %

The lexicographic database at Birzeit University \cite{JA19} provides a large set of MSA lemmas, word forms, and morphological features, which are linked with the Arabic Ontology \cite{J21} using the W3C LEMON model \cite{JAM19}.

\subsection{Dialectal Arabic Resources} 

There are several Arabic dialectal corpora with diverse morphological annotations.

An early pilot to build a Levantine Arabic Tree bank is presented in ~\cite{maamouri-etal-2006-developing}. The Palestinian dialect corpus Curras ~\cite{EJHZ22,JHRAZ17,JHAZ14} comprises about $56K$ tokens. Each word in the Curras was annotated with different morphological features, including Prefixes, Stem, Suffixes, MSA lemma, Dialect Lemma, Gloss, POS, Gender, Number, and Aspect. The Lebanese Baladi corpus ($9.6K$ tokens) was developed in the same manner as Curras in order to form a more Levantine corpus \cite{EJHZ22}. 

CALLHOME~\cite{CallHome_1997} is an Egyptian Arabic corpus with transcripts of telephone conversations in Egyptian. CALIMA~\cite{maamouri-etal-2006-developing} extended ECAL~\cite{EcalKilany_2002} which built on CALLHOME to provide morphological analysis of Egyptian. The COLABA project~\cite{Colaba_diab_2010} collected Egyptian and Levantine resources from online blogs leading to the construction of Egyptian Tree Bank (ARZATB)~\cite{maamouri-etal-2014-developing}.

The Lisan~\cite{jarrar2022lisan} consists of $1.2$ million tokens, covering Iraqi, Yemeni, Sudanese, and Libyan dialects. The Yemeni corpus (about $1.05$M tokens) was collected automatically from Twitter, while the other three dialects (about $50$K tokens each) were manually collected from Facebook and YouTube. Each word in the four corpora was annotated with different morphological features, such as POS, stem, prefixes, suffixes, lemma, and a gloss in English.  

A corpus of $200K$ tokens was morphologically annotated covering seven different Arabic dialects including Taizi, Sanaani, Najdi, Jordanian, Syrian, Iraqi, and Moroccan ~\cite{SevenCorpora}. The GUMAR Emirati corpus~\cite{khalifa-etal-2018-morphologically} consists of $200$K tokens collected from novels. MADAR~\cite{bouamor-etal-2018-madar} is an ongoing multi-dialect corpus covering $26$ cities and their corresponding dialects. The Arabizi Tunisian corpus has $42$K tokens ~\cite{gugliotta-dinarelli-2022-tarc}.

The NADI (nuanced Arabic dialect identification) SharedTask~\cite{abdul-mageed-etal-2021-nadi,abdul-mageed-etal-2020-nadi} provided researchers with $10$-million/$21$K unlabeled/labeled tweets and challenged researchers to identify the province-level dialects across $21$ countries.

%% file: nabra_s3_syrian.tex
The Levantine family of dialects can be linguistically split across the north including  Lebanon and Syria, and the south including Palestine and Jordan. During the seventh century, Arabic spread across the area, which spoke Western Aramaic before then~\cite{skaf:tel-01368247}. 

Aramaic is a Semitic language continuum spoken during antiquity throughout the Levant where  It served as the {\it lingua-franca}. 
Aramaic survives today through modern dialects such as Turoyo Syriac and Western Neo-Aramaic spoken in parts of Syria. 
It also survives more subtly in the noticeable substratum underlying Levantine dialects that differ from MSA on several linguistic characteristics such as phonology, syntax, morphology, and lexicon. 
This additionally motivates the development of morphologically annotated resources for Levantine dialects. 
In the sequel, we briefly review the differentiating factors between Levantine dialects, Syrian dialects, and MSA.

\subsection{Levantine Phonology}
Aramaic variants use the Abjad alphabet composed of 22 letters. 
When Arabic spread, the population of the region transcribed Arabic with its 28 letters using the 22-letter Abjad resulting in ``Garshouni'', a Syriac writing tradition~\cite{briquelchatonnet:hal-00278273}. 
Adaptations to fit the additional letters led some Syriac graphemes to represent multiple phonemes of Arabic, especially some of the emphatic letters.

\subsection{Syrian Phonology and Orthography}
The Syrian Dialect has a glottal stop phoneme  /\textglotstop/~
that is cognate with either  
Hamza \Ar{إ أ ؤ ئ}\TrAr{ء} or  Qaf \TrAr{ق}. 
In spontaneous Syrian orthography,
the two forms are distinguished in a manner 
similar to Lisan guidelines~\cite{JZHNW23}. 
Exceptions include  
\TrAr{هلأ}(now) written \TrAr{هلق} in $Token$ with normalization rules to highlight its etymology link to \TrAr{هالوقت}(this time). 
Less common spelling variations include devoicing \TrAr{ج} ~/\textyogh/ ~
to /\textesh/,
which sometimes  reflects in spontaneous orthography, e.g., \TrAr{نجتمع}   
/ni\textyogh tmi\textglotstop/ 
(we meet) may appear as \TrAr{نشتمع} 
/ni\textesh tim\textrevglotstop/.

\subsection{Levantine Morphology}
Levantine inherits templatic morphology from Semitic languages where affixes play important roles. Several morphological differences exist when compared to MSA.

\begin{itemize}[leftmargin=.5em]
\item Diacritic marking for syntax roles is less required in Levantine. 
They are marked with suffixes resulting 
in similar phonetic effects. 
For example, there is no need for writing
Dhamma \TrAr{ـُ} to distinguish the subject from the object. 
The MSA sentence \TrAr{غلب البطلُ الأسدَ} (The hero conquered the lion) may switch the subject and object as in \TrAr{غلب الأسدَ البطلُ} and the diacritics
distinguish the roles. 
The Levantine variants are \TrAr{البطل غلب الأسد}
and \TrAr{الأسد غلبو البطل} 
(also written as \TrAr{الأسد غلبه البطل}) 
with no need for diacritics. 
\item Some Levantine-specific morphemes  do not exist in MSA such as \TrAr{عم}
which denotes present continuous tense when it precedes imperfect verbs
\TrAr{أنا عم باكل } (I am eating). 
Without it
\TrAr{أنا باكل}
means the general truth (I eat). 
MSA lacks such an indicator and the tense is
inferred from context: \TrAr{أنا آكل} can mean both "I am eating'' or "I eat''. 

\item Other morphemes include \TrAr{رح} and \TrAr{ح} that are  Levantine future indicators compared to  MSA’s \TrAr{س} and \TrAr{سوف}. 
(iv) The progressive Levantine particle \TrAr{بـ } (as in \TrAr{باكل})  indicates imperfective verbs and no counterpart exists in MSA. 
\end{itemize}



Syrian dialects lack the  negation enclitic 
\TrAr{ش} in a distinction from southern Levantine dialects. 
Syrian dialects make use of a number of future particles in free distribution. 
The progressive particle \TrAr{عم}  strictly indicates active momentarily progression,
while the progressive proclitic +\TrAr{ب} indicates a wider habitual to the progressive range.

\subsection{Levantine Dialect Lexicon} 
The Levantine lexicon is rich with  
loan words from other languages due to its 
cross-civilization frequent passage location.


Some Syrian words are  originally Syriac, e.g., 
\TrAr{شوب} (hot), or \TrAr{براني} (outer). 
Other words are originally Turkish, e.g., \TrAr{دغري} (straightforward). 
Some words encountered major semantic shifts, e.g., \TrAr{طز} comes from Turkish tuz for ‘salt’, 
then semantically shifted to mean `something unimportant', and eventually `good riddance'. 
Other words were  borrowed from French, e.g., \TrAr{ديكور} (decor) and \TrAr{جاتو} (gateaux), 
and from Persian, e.g., \TrAr{سرسري} (badman).
Military terms \TrAr{كورنيت} are used to specify accuracy and sharpness.  

%% file: nabra_s4_local.tex
Syrian Arabic dialects are used in daily communication among most Syrians. 
Some of them are closer to Iraqi dialects, and the rest are closer to the Levantine southern Levantine dialects.  
Here, we review the most famous dialects spoken in Syria. 

{\bf The Shami dialect~} 
is the dominant dialect in the Damascus area and is the most widespread and used Syrian dialect.
As the dialect of the capital, it dominates Syrian series and films which are widely accepted, appreciated, and spread in the Arab world. 
It is used in dubbing and translation of foreign series (Turkish and Hindi). 

Table~\ref{t:shami_examples}
 shows Shami dialect features:

\begin{itemize}[leftmargin=.5em]
\item
Sculpture: abbreviate two or more words.
\item
Substitution: an example is the replacement of  \TrAr{ق} with \TrAr{ء} hamza.
\item
Spatial inversion: the introduction or delay of letters to simplify pronunciation.
\item
Inclination: vowel exchange where \TrAr{ا} 
is pronounced \TrAr{ي}.
\end{itemize}

\begin{table}
    \resizebox{.5\textwidth}{!}{
    \begin{tabular}
    {lccc}
   Shami & MSA	& Gloss	& Rule \\ \hline
   \Ar{شو بدّك} & \Ar{أي شيء بودّك} & what do  & \Ar{النحت} \\ 
   \TrNoAr{شو بدّك} & \TrNoAr{أي شيء بودّك} &  you want? &    Sculpture \\ \hline 
   
   \Ar{بالمشرمحي} & \Ar{بكلام عربي واضح وفصيح} &In clear & \Ar{النحت} \\ 
   \TrNoAr{بالمشرمحي} & \TrNoAr{بكلام عربي واضح وفصيح} & words & Sculpture \\ 
   \hline
   
   \Ar{أديش} & \Ar{كم يساوي} & how much & \Ar{ابدال} \\ 
   \TrNoAr{أديش} & \TrNoAr{كم يساوي} &  & Substitution \\ 
   \hline
   
   \Ar{جوز} & \Ar{زَوْج} & husband & \Ar{قلب المكاني}\\ 
   \TrNoAr{جوز} & \TrNoAr{زَوْج} &  & spatial inversion\\ 
   \hline

   \Ar{هنيك} & \Ar{هناك} & There & \Ar{إمالة} \\ 
   \TrNoAr{هنيك} & \TrNoAr{هناك} & & inclination \\ \hline
    \end{tabular}
    }
    \caption{\label{t:shami_examples} Examples of Shami Dialect}\vspace{-1.5em}
\end{table}

{\bf The Aleppo dialect~}
is dominant in Aleppo in northern Syria.
It is distinctive in pronunciation and has a unique vocabulary used in Aleppo alone. The distinct vocabulary comes from ancient Syriac or Turkish. 
Examples of Syriac and Turkish vocabulary used in Aleppo follow. 
Syriac 
\TrAr{إيمت }
replaces MSA  
\TrAr{متى}
(when), and Syriac 
\TrAr{دعك}
replaces MSA  \TrAr{عجن}
(knead). 
Turkish \TrAr{فرتيكة } and \TrAr{سكرتون} replace MSA \TrAr{شوكة} (fork),   \TrAr{خزانة} (closet), respectively.

With non-Arabic Syriac vowels (e, o), Aleppo words and verbs do not need  
the Dammah \Ar{ـُ} (nourishing) and fatha \Ar{ـَ} (accusative) diacritics. 
Verbs may require more than one object denoting the concept of \TrAr{تعدي} (exceeds).
Verbs connect to \Ar{ن} to denote the masculine plural instead of the MSA 
suffix \TrAr{م} 
Turkish influence on Aleppo dialects morphs the pronunciation of fixed letters such as \TrAr{ج}  and \TrAr{ق} to a majestic Turkish tone, and also reduces the pronunciation of vowels. 

{\bf The Latakia dialect~}
is spoken across the coast in Latakia and Tartous. 
It is a mixture of Arabic, Syriac, and  Phoenician. 
It is characterized by the strong pronunciation of the letter \TrAr{ق}, and also features the letter \TrAr{م} before verbs to denote the present tense in all its forms, e.g.\TrAr{منكتب} (we write/are writing), \TrAr{ميدرس} (he studies/is studying).

{\bf The Raqqa dialect~}
is one of the closest dialects to classical Arabic in terms of vocabulary.
Raqqa enjoys a distinguished location on the shores of the Euphrates River. 
It is home (\TrAr{ديار}) Mudar, who are Arabs from the north.
Mudar were displaced to the Euphrates island several centuries before Islam. 
The Raqqa syllables sound commensurate to the corresponding classical Arabic syllables. For example,  the pronunciation of \TrAr{كـ} 
results in a thirsty \TrAr{ج} as in  \TrAr{كانت} pronounced as \TrAr{جانت}. 
The letter \TrAr{ق} is pronounced \TrAr{كـ} similar to Yemeni dialects
as in \TrAr{قاع} (earth)
pronounced as 
\TrAr{كاع}.

{\bf The Deir-ezzur dialect ~} 
aka. as \TrAr{الديرية} is in proximity to the Euphrates as well, and preserves most of the phonetic aspects of standard Arabic. 
The significantly different phonemes are \TrAr{ق}, \TrAr{ك} and \TrAr{ء}, 
while there is no 
different in the gingival sounds. 

{\bf The Homs dialect~}
varies slightly across several rural and urban areas in the Homs district. 
This is mainly due to the habitual diversity of the countryside including 
a sizeable Turkman population. 
This paper covers the dominant variant in the city of Homs. 
The Homs dialect is characterized by pronouncing the first letter in a word
as if it has a Dammah 
\TrAr{ـُ} diacritic (inclusion).
This includes the name of the city \TrAr{حِمص},
 pronounced with a Kasra \TrAr{ِ} dialect everywhere else. 
It also flips gender when it comes to masculine second-person
\TrAr{إنتِ} (you-male in Homsi)
and feminine second person 
\TrAr{إنتَ} (you-female in Homsi). 
It also differs in the pronunciation of the
letter \TrAr{ج} as they phonetically
annex a silent \TrAr{د} resulting in a \TrAr{دج}
sound.

{\bf The Hama dialect~}
is spoken in the central Syrian governorates. 
It is a good representative of the Syrian Levantine
dialects and close to the Shami one, as it tends to be soft and long in speech. 
It is distinguished by its eloquence and stretch in speech. 
Al-Hader (city in Hama) variant of the Hama dialect is the most prominent variant. 

{\bf The Hauran dialect~}
is spoken south of the Damascus countryside down to the Ajloun mountains in Jordan
including Daraa. 
It is an ancient Arabic dialect spoken by multiple Arab tribes, where each of them has some distinguishing phonetic characteristics.

{\bf The Al-Suwayda dialect~} 
is spoken in Jabal al-Arab.
The harshness of the mountain environment is reflected in the dialect's tone.
It is taut,  clear, and possesses a fast rhythm. 
Syllable notes exit soundly and eloquently.
The concept of \TrAr{المضافة} 
played a major role in preserving the strength 
of the dialect. 
Therein, prominent, cultured, and experienced speakers exchange arguments. 
This highly contributed to the rigor of the dialect and brought it closer to standard and classical Arabic. 

{\bf The  Mardini dialect~}
takes its name from the city of Mardin in \TrAr{الحسكة}.
It is also called \TrAr{الجزراوية} in relevance to the \TrAr{الفراتية} island.
The dialect contains many Turkish, Persian, and Aramaic words.

%% file: nabra_s5_collection.tex
We manually collected about 6,000 sentences with 60K tokens from Facebook, blogs, popular proverbs, Syrian films and series, local poetry, and lyrics of 
popular local songs in several Syrian dialects to build \nabra. 
Table~\ref{t:local_dialect_stats} provides statistics on tokens, unique tokens, sentences, lemmas, nouns, verbs, and functional words in each of the 10  dialects \nabra covers.

The distribution relatively follows the order of  dialect demographics. 
The Shami dialect is the richest with 17.3K tokens, used as  primary dialect in Damascus, the capital, and in various Syrian TV series and films.
\nabra contains 9.2K Aleppo tokens  collected from popular stories on Facebook and from vocal poetry. 
Coastal Latakia  features 7.9K tokens collected from film dialogues such as \TrAr{رسايل شفهية}-\TrAr{قمران وزيتونة} (Voice letters, Qumran and Zeitouna) 
and series such as \TrAr{ضيعة ضايعة} (lost town).
We also added common proverbs.
Suwayda dialect features 3.2K tokens from the \TrAr{الخربة} series. 
For Homs and Hama we collected jokes, and food discussions from social media blogs. 
 
The Raqqa, Huran, and Mardin dialects feature the remaining 6.3K, 3.8K, and 1.6K tokens, respectively. 
We manually collected texts from social media for Raqqa and Huran. We found blogs documenting Raqqa. 
We used blogs and traditional stories for Raqqa, vocal poetry and lyrics of popular folklore songs for Mardini, 
and scenes from the Bedouin series for Huran dialects. 
We noticed that the collected data reflected  spontaneous  dialect documentation all across, 
contrary to what one would expect. Films and series were no less spontaneous than blogs and social media.  

As Arabic is diacritic-sensitive \cite{JZAA18}, we did not remove any diacritics We tokenized the text of \nabra so that each token has a tuple with the following information. 

$\langle$SentenceID, TokenID, TokenText, LocalDialectName, Governate$\rangle$

\begin{table*}[!ht]
\small
    \centering
    \resizebox{.98\textwidth}{!}{
    \begin{tabular}{|p{2.7cm}|p{1.3cm}|p{1cm}|p{1cm}|p{1cm}|p{1.1cm}|p{0.8cm}|p{1cm}|p{1.4cm}|p{0.8cm}|p{1cm}|}
    \hline
        Dialect \Ar{لهجة} &
        Damascus (Shami) \Ar{شامية} &
        Aleppo \Ar{حلبية} &
        Latakia  \Ar{ساحلية}  &
        Raqqa  \Ar{رقاوية} &
        Deir-Ezzur  \Ar{ديرية} & 
        Homs \Ar{حمصية}  & 
        Huran \Ar{حوران} & 
        Suwayda \Ar{سويداء} & 
        Hama \Ar{حموية} & 
        Mardin \Ar{ماردلية}  \\     \hline
        Tokens & \makecell[r]{17,274} & \makecell[r]{9,255} & \makecell[r]{7,893} & \makecell[r]{6,284} & \makecell[r]{4,322} & \makecell[r]{4,139} & \makecell[r]{3,807} & \makecell[r]{3,150} & \makecell[r]{2,322} & \makecell[r]{1,575} \\ \hline
        Unique Tokens & \makecell[r]{7,123} & \makecell[r]{4,452} & \makecell[r]{3,829} & \makecell[r]{3,389} & \makecell[r]{2,453} & \makecell[r]{2,047} & \makecell[r]{2,094} & \makecell[r]{1,681} & \makecell[r]{1,355} & \makecell[r]{949} \\ \hline
        Sentences & \makecell[r]{1,181} & \makecell[r]{787} & \makecell[r]{829} & \makecell[r]{679} & \makecell[r]{519} & \makecell[r]{518} & \makecell[r]{457} & \makecell[r]{381} & \makecell[r]{340} & \makecell[r]{243} \\ \hline
        Unique MSA Lemma & \makecell[r]{4,230} & \makecell[r]{2,825} & \makecell[r]{2,548} & \makecell[r]{2,367} & \makecell[r]{1,909} & \makecell[r]{1,543} & \makecell[r]{1,580} & \makecell[r]{1,312} & \makecell[r]{1,051} & \makecell[r]{686} \\ \hline
        Unique DA lemma & \makecell[r]{4,351} & \makecell[r]{2,969} & \makecell[r]{2,681} & \makecell[r]{2,490} & \makecell[r]{1,954} & \makecell[r]{1,591} & \makecell[r]{1,646} & \makecell[r]{1,354} & \makecell[r]{1,095} & \makecell[r]{710} \\ \hline
        Nouns & \makecell[r]{7,700} & \makecell[r]{4,251} & \makecell[r]{3,771} & \makecell[r]{3,316} & \makecell[r]{2,384} & \makecell[r]{2,064} & \makecell[r]{2,090} & \makecell[r]{1,527} & \makecell[r]{1,135} & \makecell[r]{694} \\ \hline
        Verbs & \makecell[r]{3,524} & \makecell[r]{1,897} & \makecell[r]{1,557} & \makecell[r]{985} & \makecell[r]{714} & \makecell[r]{709} & \makecell[r]{518} & \makecell[r]{554} & \makecell[r]{369} & \makecell[r]{339} \\ \hline
        Functional Words & \makecell[r]{6,027} & \makecell[r]{3,090} & \makecell[r]{2,560} & \makecell[r]{1,960} & \makecell[r]{1,213} & \makecell[r]{1,359} & \makecell[r]{1,194} & \makecell[r]{1,069} & \makecell[r]{815} & \makecell[r]{534} \\ \hline
    \end{tabular}
    }
    \caption{\label{t:local_dialect_stats} Counts of tokens, unique tokens, sentences, unique MSA lemmas, unique dialectal lemmas, Nouns, Verbs, and functional words for each of the Syrian dialects}
\end{table*}

%% file: nabra_s6_annotation.tex
We followed a semi-automated methodology, with an integrated productivity tool, friendly to non-programmers, to annotate \nabra. 

\subsection{Methodology} 
\begin{figure*}[tb]
    \centering
    \includegraphics[width=1\textwidth]{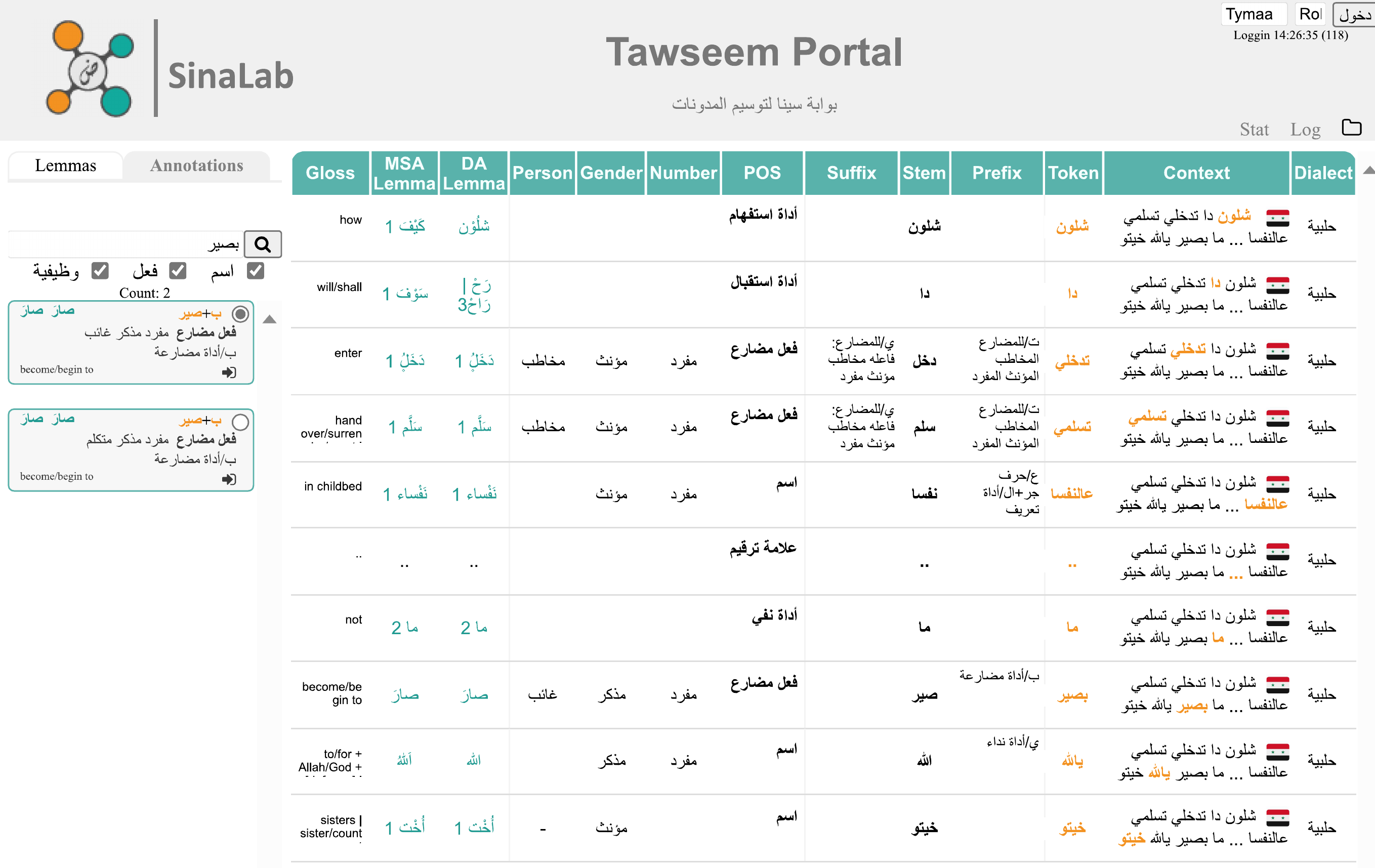}
    \caption{Screenshot of the \textit{Tawseem} annotation portal, our web-based annotation tool}
    \label{fig:Sense_annotated}
\end{figure*}

We developed the \textit{Tawseem} annotation portal 
to help automate and validate the annotation process. The portal leverages spreadsheets, familiar to common users, and is powered by smart functionalities to improve
annotation productivity.
Figure~\ref{fig:Sense_annotated} shows  a snapshot of
\textit{Tawseem} annotation portal with the sentence 
\TrAr{شلون دا تدخلي تسلمي عالنفسا} (how would you enter to greet someone in childbed).  

For each token in the sentence, the portal saves 17 data elements.
The $SentenceID$ and $TokenID$ columns identify the sentence and token.
The rest of the columns specify the $rowToken$, $Token$, $prefix(s)$, $stem$, $suffix(s)$, $POS$, $gender$, $number$, $person$, $aspect$, $MSA lemma$, $dialect lemma$, $synonym(s)$, $gloss$, as well as the $sub-dialect$.

To simplify and accelerate the annotation process we leverage existing annotations in the following manner.
First, we uploaded existing annotated corpora for dialects and MSA ~\cite{EJHZ22,JZHNW23} into the $Tawseem$ tools. 

The tool allows the annotators to search and look up previous annotations.
The lookup services search the database 
and return the top matching results ranked. 
Annotators can then select one of the results, 
and correct the corresponding features if needed. 

Second, annotators can search the \textit{Tawseem} portal annotations in other sentences whether made by themselves or by other annotators. 
This helps leverage previous annotations and improves the correction process. Additionally, annotators can look for existing annotations of a specific token in the \textit{Tawseem} portal results. 

\subsection{Annotation Guidelines}

Training annotators to use the \textit{Tawseem} portal was straightforward as they were all familiar with the interface of a productivity spreadsheet. We also trained them with annotation guidelines for each of the features in \nabra as follows:  

{\bf rowToken}: $rawToken$ is the raw word as it appears in the corpus, without any modification.

{\bf Token~}: $Token$ is the normalized
version of the $rawToken$. This entry corrects spelling errors if needed. The idea is to unify different forms of spelling the same word with one specification to mitigate the lack of spelling rules for Arabic dialects.
It is necessary to unify the different ways one word can be written by multiple users 
to reflect the same pronunciation. 
We adopted the $Token$ guidelines used in the Lisan corpora \cite{JZHNW23} as well as the Palestinian Curras2 and Lebanese Baladi corpora \cite{EJHZ22}
so that \nabra can be included smoothly in a larger family of Arabic dialects 
for further research and applications if needed. 


{\bf Dialect lemma~} 
(\Ar{المدخلة المعجمية العامية})
determines the dialect's original source of the token.
Thus, if the word is a verb, we choose the past masculine 3rd person singular form as its colloquial origin. 
For nouns, we select the singular masculine, if not attained we select the singular feminine form.
When introducing a new lemma, we specify the following:
(i) definitions of senses in Arabic, which is important for word sense disambiguation tasks \cite{HJ21b,JMHK23} and Word-in-Context WiC disambiguation tasks \cite{HJ21}. 
(ii) Equivalent lemmas in MSA \cite{JAM19,JKKS21}.

{\bf MSA Lemma~} 
(\Ar{المدخلة المعجمية الفصحى})
determines the MSA original source of the token. 
Table~\ref{t:annotation_examples} shows examples of some tokens with their $Token$, and dialect and MSA lemmas. 
 
The $Tawseem$ portal allows to search for lemmas in the Birzeit’s Lexicographic database~\cite{JA19,ADJ19} and Arabic
Ontology~\cite{J21,J11}; otherwise, we introduced a new lemma.

{\bf The Synonym} (\Ar{المرادف})
feature provides synonyms for the token and sometimes explains the token semantics. We used an online tool for automatic synonym discovery \cite{GJJB23,KAEMKRTM23}.

{\bf Gloss} (\Ar{المعنى بالانجليزية}) specifies the meaning of the token in English.
It typically specifies a short definition of lemma semantics. See an elaboration on the gloss formulation guidelines in \cite{J06}.

{\bf POS} (\Ar{قسم الكلام})
specifies the part of speech of the token. 
This concerns the grammatical category of the token.
We follow the SAMA tagset for compatibility reasons
\cite{MaamouriSama2010}.

{\bf Stem~} (\Ar{الجذر})
specifies the segment of the token after removing suffixes and prefixes. 
It helps in the morphological analysis of the tokens. 
We follow the (Stem/POS) tagging schema used in~\cite{MaamouriSama2010} 
where the stem and POS are specified separated by ’/’.

{\bf Affixes: prefixes and suffixes.~} 
We follow the prefixes 
\Ar{السوابق} and suffixes \Ar{اللواحق} tagging schema used in SAMA.  

⟨Prefix1/POS⟩ + ⟨Prefix2/POS⟩ $\ldots$ 

⟨Suffix1/POS⟩ + ⟨Suffix2/POS⟩ $\ldots$ 

The schema specifies a sequence of affix and affix POS pairs separated by '+'.
Each pair is an affix and affix POS separated by '/'.

Affixes and stems are morphemes where the concept of morpheme denotes the smallest 
morphological unit of text. 
Prefixes specify morphemes that connect to the beginning of a stem or to other prefixes
to form a word. 
Suffixes specify morphemes that connect to the end of a stem or to other morphemes to form a word. 
Dialect affixes and their POS tags differ from MSA affixes and augment them 
due to the extended morpho-syntactic and semantic roles of dialect affixes.

Note here, for example, the synergy of using the future and progressive particles
\Ar{ع/أداة  استقبال} (FUT\_PART) + \Ar{ ب/أداة مضارعة }(PROG\_PART) 
as prefixes to indicate present continuous tense for verbs in Aleppo as in 
\TrAr{عبشتغل} 
(I am working). 

While most of the Syrian dialects precede present tense verbs with the IV1P POS with
\Ar{م/أداة مضارعة } (PROG\_PART),
the Latakia coastal dialect applies it to almost all present tense verbs as with
\TrAr{مأدرس}  (I am studying). 
Latakia dialect also uses the prefix \TrAr{أ} for negation (and thus it corresponds to a 
NEG\_PART POS tag) before present tense verbs as in \TrAr{أبعرف} (I don’t know). 


{\bf Person ~} (\Ar{ الإسناد})
specifies whether the subject of the token is a \TrAr{متكلم} (first), 
(\TrAr{مخاطب}) (second) or  \TrAr{ غائب} (absent) person when applicable. 

{\bf Aspect} (\Ar{صيغة الفعل}) 
concerns verbs and specifies whether they are  in 
(\TrAr{مضارع}) present for imperfective verbs 
(\TrAr{ماضي}) past for perfective verbs 
and (\TrAr{أمر}) imperative tense. 

{\bf Gender } (\Ar{الجنس}) specifies whether a word is 
of \TrAr{مذكر}  male for masculine, 
\TrAr{مؤنث} female for feminine, or
\TrAr{لا ينطبق}  not applicable association when applicable. 

{\bf Number} (\Ar{العدد})
denotes \TrAr{مفرد} for singular, \TrAr{جمع} for plural, \TrAr{مثنى} 
for dual (to count two units), or \Ar{لا ينطبق} for uncountable words when applicable.

\begin{table}
\resizebox{.45\textwidth}{!}{
\begin{tabular}{llccc}
$rowToken$  &   & $Token$ & Dialect lemma & MSA lemma \\ \hline
\TrAr{ألت}& I said & \TrAr{قلت} & \TrAr{قال} & \TrAr{قَالَ} \\ \hline

\TrAr{تختك}& your bed & \TrAr{تختك} & \TrAr{تخت} & \TrAr{سَريْر} \\ \hline

\TrAr{مهندز} & engineer & \TrAr{مهندس} & \TrAr{مهندس} & \TrAr{مُهَنْدِس} \\ \hline
\TrAr{طريئ} & street & \TrAr{طريق} & \TrAr{طريق} & \TrAr{طَرِيق}
\end{tabular}
}
\caption{\label{t:feature_examples} Example annotations for \nabra tokens
\label{t:annotation_examples}}
\end{table}

%% file: nabra_s7_evaluation.tex
\input{iaa_tables.tex}
 Before evaluating \nabra, we normalized the annotations to unify variant annotations that are equivalent. 
These variants occur due to human mistakes such as typos
(\TrAr{ماصي} instead of \TrAr{ماضي}), 
ordering of tags in sequences of tags, 
and inconsistent use of separators and spacing.

Another source of variants is tokens with no feature values 
in the existing annotated dialects. 
Annotators have to come up
with novel values. 
We detected these tag values, ranked them based on their frequencies, and clustered them based on their edit distance from each other. 
Then we reviewed them and unified them across \nabra and its features. 

We developed a small suite of VBA scripts
empowered with regular expressions 
to check for these variants
and correct them automatically where possible.
If automatic correction is not possible and human attention is required, then our reference annotators interfere to correct it.

\subsection{Inter-annotation agreement} 

After the automatic corrections, six linguists visited the annotations to approve or correct them. 
This created a significant overlap of annotations as shown in Table~\ref{t:iaa_overlap}. 
The overlap column shows the number of annotations per feature that 
had more than one annotation. 
Some of the second annotations were performed by the original annotator, 
so the reviewed column shows the number of annotations
that were reviewed by two or more annotators. 
The unique column shows the number of unique values
for the tokens with overlapping annotations. 

The correction approach secured a significant overlap. 
We report the performance of the annotators in terms of precision,
recall, and F1-score taking the correcting annotator as a 
reference in Table~\ref{t:iaa_prfk}. 
A true positive (TP) for a feature value $fv$, 
denotes that the original annotation matched the 
reference annotation. 
A false positive (FP) for $fv$ reflects an original annotator selecting
$fv$ for the token in conflict with the selection of the reference 
annotator.
A false negative (FN) is when the original annotator fails to select 
$fv$ for a token when the reference annotator selected it. 
Precision (P) and recall (R) are given by the ratios $TP/(TP+FP)$, and $TP/(TP+FN)$, 
respectively. 
The F1-score is given by $ 2 P R/(P+R)$. 

We also computed the Kappa-Cohen metric~\cite{j:kappa:McHugh2015}
as implemented in the Scientific Kit Learn package~\cite{w-scikit-learn}.
Table~\ref{t:iaa_prfk}  
shows the results where we compared the feature values of 
the reference annotators versus those of the original annotators. 

The results show performance and agreement across 
all features. 
The $\kappa$ scores are lower than the F-scores 
as the the $\kappa$ metric accommodates 
for agreement by chance.
The difference shows more with prefixes and 
suffixes as a significant part of the tokens had
empty prefix and suffix, allowing more agreement by chance.

%% file: iaa_tables.tex
\begin{table*}[tb]
\centering
\begin{tabular}{c||c|c|c||c|c|c||c} 
\hline 
Feature&TP&FP&FN&P&R&F&$\kappa$
\\ \hline \hline
Stem&21,506&4,933&5,461&0.813&0.797&0.805&0.796
\\ \hline
POS&20,727&2,979&3,316&0.874&0.862&0.868&0.843
\\ \hline
Prefix&22,886&448&496&0.981&0.979&0.980&0.939
\\ \hline
Suffix&22,096&1,247&1,380&0.947&0.941&0.944&0.837 \\ \hline
DA Lemma&18,600&5,765&6,451&0.763&0.742&0.753&0.739 \\ \hline
MSA Lemma& 19,300&5,161&5,749&0.789&0.770&0.780&0.767\\ \hline
\end{tabular}
\caption{Precision and recall results due to annotation correction with $F$ and $\kappa$ scores \label{t:iaa_prfk}}
\end{table*}

\begin{table}[tb]
\begin{tabular}{c||c|c|c} \hline
Feature & Overlap & Reviewed  & Unique \\ \hline \hline
Stem& 44,687 & 26,967  & 3,102 \\ \hline
POS & 39,007 & 24,043  & 56 \\ \hline
Prefix & 39,007 & 23382  & 163 \\ \hline
Suffix & 39,007 & 23,476 & 358  \\ \hline
DALemma & 41,579 & 25,052 & 3,586  \\
\hline
MSALemma & 41,579 & 25,050 & 3,352   \\
\hline
\end{tabular}
\caption{Reviewed overlap and unique feature values across \nabra \label{t:iaa_overlap}}
\end{table}

%% file: nabra_s8_qualitative_evaluation.tex
To conduct a qualitative evaluation, we randomly selected about $7K$ annotations and reviewed them manually.
We found a high agreement between the annotators who followed the specific guidelines and used our annotation tool. 
In what follows, we discuss some of the common mistakes:

\begin{itemize}
\item[(i)] In rare cases, tokens specific to small local communities were hard to understand, Such as the token \TrAr{زنطر} (become cold) in the Latakia dialect.
Although the annotators did their best to search external resources to understand such words, some mistakes still existed.

\item[(ii)] Tokens with no clear MSA equivalent led to difficulty in selecting MSA lemmas; thus, different annotators might not agree on selecting the same lemma. For example, the token \TrAr{عَمنوَّل} may have several MSA lemmas, such as \TrAr{عام} (year), or \TrAr{ماضي} (past).

\item[(iii)] Semantic ambiguities in contexts led to disagreements on selecting lemmas. 
For instance, the token \TrAr{بقى} has three possible meanings (was), (therefore) and (also). And sometimes all three fit the context. 

\end{itemize}

%% file: nabra_limitations.tex
The work in \nabra has the following limitations. 
\begin{itemize}
    \item \nabra covers $10$ Syrian dialects. variants of these dialects and other smaller dialects confined in less urban localities exist. 
    Future work should extend \nabra to better cover the Syrian dialect.
    \item \nabra addressed the Syrian dialects and their relation to the Arabic language and touched in prose on the relations to languages of origin such as Aramaic and Cyrillic. 
    More data-oriented work is needed to relate \nabra to languages of origin that were spoken in Syria as well as to the geo-linguistic features of these languages. 
    \item The annotation and evaluation process leveraged linguists who may be
    better at some of the dialects than others. We will make \nabra available online with correction suggestion capacities to accommodate for possible potential corrections.
\end{itemize}
